\documentclass{article}
\usepackage[top=1.0in, bottom=1in, left=1in, right=1in]{geometry}

\usepackage{amsmath,amssymb,amstext} % Lots of math symbols and environments
\usepackage{graphicx}
\usepackage{caption}
\usepackage{subcaption}
\usepackage{multirow}
\usepackage{array}
\usepackage{float}  % for [H]
\usepackage{xspace}  % for space after macro
\usepackage{enumitem}  % custom enumerate labels
\usepackage{authblk}

\newcommand{\apriori}{\textit{a priori}\xspace}

\renewcommand{\vec}[1]{\mathbf{#1}}

\DeclareMathOperator*{\argmin}{arg\,min}
\DeclareMathOperator*{\argmax}{arg\,max}

\begin{document}

\title{A spectral-spatial fusion model for robust blood pulse waveform extraction in photoplethysmographic imaging}
%\date{}

\author[a,*]{Robert~Amelard}
\author[a]{David~A~Clausi}
\author[a]{Alexander~Wong}
\affil[a]{University of Waterloo, Department of Systems Design Engineering, 200 University Ave W, Waterloo, Canada, N2L 3G1}
%\email{$^*$ramelard@uwaterloo.ca}

\maketitle

\begin{abstract}
Photoplethysmographic imaging is a camera-based solution for non-contact cardiovascular monitoring from a distance. This technology enables monitoring in situations where contact-based devices may be problematic or infeasible, such as ambulatory, sleep, and multi-individual monitoring. However, extracting the blood pulse waveform signal is challenging due to the unknown mixture of relevant (pulsatile) and irrelevant pixels in the scene. Here, we design and implement a signal fusion framework, FusionPPG, for extracting a blood pulse waveform signal with strong temporal fidelity from a scene without requiring anatomical priors (e.g., facial tracking). The extraction problem is posed as a Bayesian least squares fusion problem, and solved using a novel probabilistic pulsatility model that incorporates both physiologically derived spectral and spatial waveform priors to identify pulsatility characteristics in the scene. Experimental results show statistically significantly improvements compared to the FaceMeanPPG method ($p<0.001$) and DistancePPG ($p<0.001$) methods. Heart rates predicted using FusionPPG correlated strongly with ground truth measurements ($r^2=0.9952$). FusionPPG was the only method able to assess cardiac arrhythmia via temporal analysis.
\end{abstract}

%\ocis{(100.2960) Image analysis; (170.3880) Medical and biological imaging; (170.1470) Blood or tissue constituent monitoring.}

% Include email contact information for corresponding author
%{\noindent \footnotesize\textbf{*}Robert Amelard,  \linkable{ramelard@uwaterloo.ca} }

%\begin{spacing}{2}   % use double spacing for rest of manuscript

\section{Introduction}
\label{sect:intro}
Photoplethysmography (PPG) is a safe and inexpensive cardiovascular monitoring technology~\cite{allen2007}. Using an illumination source and detector, transient fluctuations in illumination intensity pertaining to localized changes in blood volume enable non-invasive probing of vascular characteristics. Traditionally, PPG is monitored via a contact probe that operates in either transmittance or reflectance mode, where the source and detector are placed on opposite or the same side of the tissue, respectively. However, this conventional type of monitoring provides hemodynamic information only for a single point, and is an impractical measurement tool in settings such as ambulatory and multi-individual monitoring.

Recent studies have focused on developing and validating photoplethysmographic imaging (PPGI) systems~\cite{sun2016}. These systems substitute contact-based detectors with a camera and use non-contact illumination sources, enabling non-contact cardiovascular monitoring from a distance~\cite{allen2014}. The additional visual context can enable functionality such as motion compensation during exercise~\cite{sun2011}, multi-individual tracking~\cite{poh2010}, and spatial perfusion analysis~\cite{kamshilin2011}. One major challenge in PPGI is the automatic extraction of a blood pulse waveform from the video. Decoupling the detector from the body causes several challenges, such as illumination variations, motion changes, and hair occlusion. Furthermore, locations on the body that contain pulsatile flow are not readily apparent.  Identifying pixels representing the skin does not guarantee pulsatile blood flow, as some areas may be minimally vascularized, or may not contain pulsatile vessels. In fact, the pulsatile nature of the blood pulse waveform is not fully understood~\cite{allen2014,kamshilin2015}.

Existing methods for automatic signal extraction broadly rely on a combination of spatial and spectral information. The RGB components in cameras with Bayer filters have been leveraged to identify the blood pulse waveform using independent component analysis~\cite{poh2010,mcduff2014}, Beer-Lambert modeling~\cite{xu2014}, and skin composition modeling~\cite{wang2015}. However, these methods rely on measuring multispectral reflectance values such as RGB, which may not be appropriate in low-light settings such as sleep monitoring. Furthermore, the tissue penetration depth of incident illumination is wavelength- and tissue-dependent~\cite{anderson1981}. To solve this problem, some methods rely only on a single wavelength (or color channel) to extract the signal through spatial analysis~\cite{sun2011,kamshilin2011,kumar2015}. Regardless of the spectrum chosen, existing methods average the pixel intensities over chosen areas, such as the facial bounding box~\cite{sun2011,poh2010,xu2014}, predefined facial areas~\cite{mcduff2014}, and facial segmentation~\cite{kumar2015}. Many of these averaging approaches evaluate success based on heart rate analysis, and do not consider temporal signal fidelity, which is important for temporal analysis such as heart rate variability. Additionally, they rely on the success of a facial tracking algorithm, which may fail due to varying lighting conditions, different face-camera perspectives, or from various facial features, and require overhead computation which may delay real-time analysis.

Here, we propose a spectral-spatial fusion method for extracting a blood pulse waveform from a set of frames from an arbitrary scene (i.e., without facial tracking). Our goal was to extract signals that exhibited both spectral and temporal fidelity, to enable both spectral and temporal analysis. Using physiologically derived \apriori spectral and spatial information about a typical blood pulse waveform, our method learns which regions contain the strongest pulsatility, and emphasizes their contribution to determine the final signal. Results across a 24-participant study show that the proposed method generated signals that exhibited significantly stronger temporal correlation and spectral entropy compared to existing methods. As a result, this method may be used in the future to assess pulsatility in different anatomical locations.

\section{Methods}
\begin{figure}
\centering
\includegraphics[width=\textwidth]{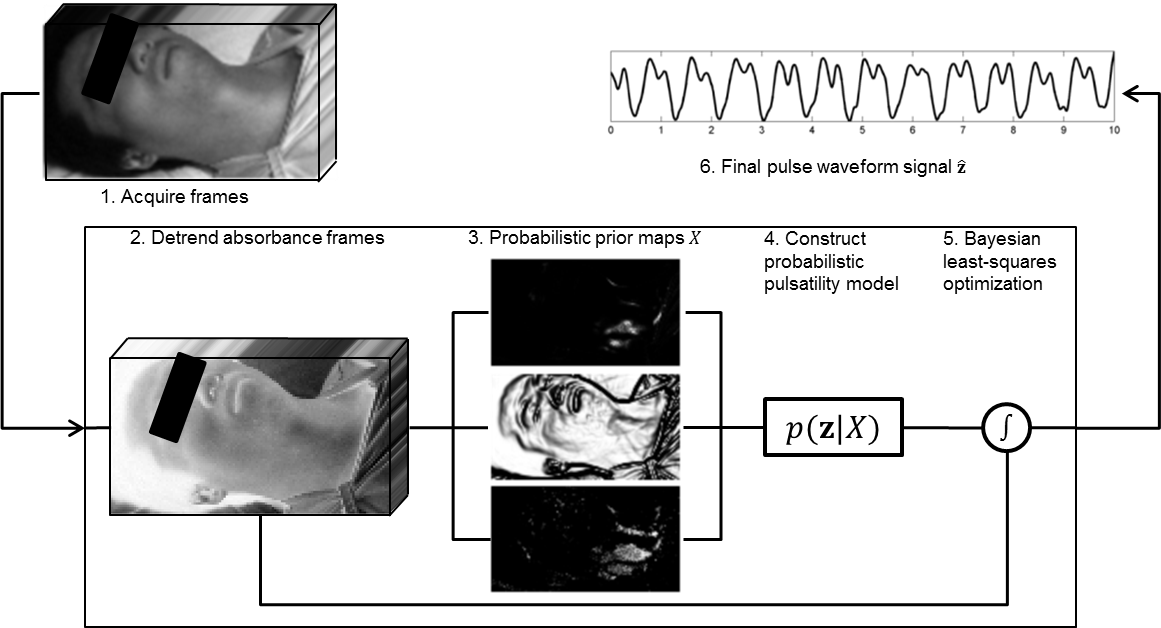}
\caption{Processing pipeline of the proposed signal extraction method. Acquired frames (1) were converted from reflectance to absorbance and detrended (2). Spectral-spatial probabilistic prior maps were computed (3) and used to model the posterior distribution representing the pulsatility model (4). Bayesian least-squares optimization was used (5) to generate the blood pulse waveform signal (6).}
\label{fig:pipeline}
\end{figure}

The goal of the spectral-spatial fusion model was to extract a clean temporal blood pulse waveform signal from a scene. By emphasizing temporal fidelity, not only can summary metrics such as heart rate be computed, but important temporal fluctuations such as cardiac arrhythmias can be assessed. The scene is assumed to contain an unknown mixture of relevant regions (i.e., skin areas which exhibit pulsatility), and irrelevant regions (e.g., background, clothing, non-pulsatile skin regions, etc.). Given this mixture of regions (input), the system must discover a temporal PPGI signal (output). Figure~\ref{fig:pipeline} provides a graphical overview of the system. Details are provided below.

\subsection{Problem Formulation}
Let $\vec{z}=z[t]$ be the (unknown) true blood pulse waveform. Let $X=\{\vec{x}_i \mid 1 \le i \le n\}$ be a set of absorbance signals, where:
\begin{equation}
\vec{x}_i=x_i(t) \cdot \sum_{k=-\infty}^{\infty} \delta(t-kT)
\end{equation}
where $\delta$ is the Dirac delta function, and $T$ is the sampling period. Here, following the Beer-Lambert law, absorbance is calculated as $x_i(t) = -\log(r_i(t))$, where $r_i(t)$ is the intensity signal for region $i$. Each signal $\vec{x}_i$ was detrended using a regularized least squares subtraction method which heavily emphasizes a smoothness prior~\cite{tarvainen2002}. Given the set of measurements $X$, which is a mixture of signals from a scene that are both relevant (e.g., skin) and irrelevant (e.g., background, skin folds, hair), the goal is to estimate the ``true'' blood pulse signal using an intelligently weighted subset of regions that contain pulsatility. This inverse problem can be formulated as a Bayesian problem, where prior physiology knowledge can be injected into the model to educate assumptions about the state (specific priors will be discussed in the following section). Mathematically, it can be solved using the Bayesian least squares formulation~\cite{fieguth2010}:
\begin{align}
  \hat{\vec{z}} &= \argmin_{\hat{\vec{z}}} \left\{ E \left[ (\hat{\vec{z}}-\vec{z})^T (\hat{\vec{z}}-\vec{z}) \mid X \right] \right\} \\
  &= \argmin_{\hat{\vec{z}}} \int (\hat{\vec{z}}-\vec{z})^T (\hat{\vec{z}}-\vec{z}) p(\vec{z}|X) d\vec{z} \label{eq:zhat2}
\end{align}
where $p(\vec{z}|X)$ is the posterior distribution of state signal $\vec{z}$ given the measurements $X$. The optimal solution is found by setting $\partial/\partial \hat{\vec{z}}=0$:
\begin{align}
  \frac{\partial}{\partial \hat{\vec{z}}} \int (\hat{\vec{z}}-\vec{z})^T (\hat{\vec{z}}-\vec{z}) p(\vec{z}|X) d\vec{z} = 0
\end{align}
Simplifying:
\begin{align}
  2 \int (\hat{\vec{z}}-\vec{z}) p(\vec{z}|X) d\vec{z} = 0 \\
  \int \hat{\vec{z}}p(\vec{z}|X) d\vec{z} - \int \vec{z}p(\vec{z}|X) d\vec{z} = 0
\end{align}
\begin{equation}
  \hat{\vec{z}} = \int \vec{z} p(\vec{z}|X) d\vec{z}
  \label{eq:bls}
\end{equation}
Thus, to solve this equation, the unknown posterior distribution $p(\vec{z}|X)$ must be modeled. This distribution represents the probability that a state signal $\vec{z}$ represents the true blood pulse waveform given the observed temporal signals $X$. The posterior distribution can be modeled as a novel probabilistic pulsatility model, which we approximated using a discrete weighted histogram of the observed states~\cite{wong2011}:
\begin{equation}
  \hat{p}(\vec{z}|X) = \frac{\sum_{i=1}^{|X|} W_i \delta(|\vec{z}-\vec{x}_i|)}{Y}
  \label{eq:whist}
\end{equation}
where $Y$ is a normalization term such that $\sum_k \hat{p}(z_k|X)=1$. The problem then becomes computing the probabilistic prior $W_i$ for each observed signal $\vec{x}_i$ to determine how well it represents the true blood pulse waveform. The following subsections propose a solution using a spectral-spatial model motivated by blood pulse waveform characteristics and vascular physiology.

\subsection{Probabilistic Pulsatility Model}
\begin{figure}
\centering
\includegraphics[width=0.8\textwidth]{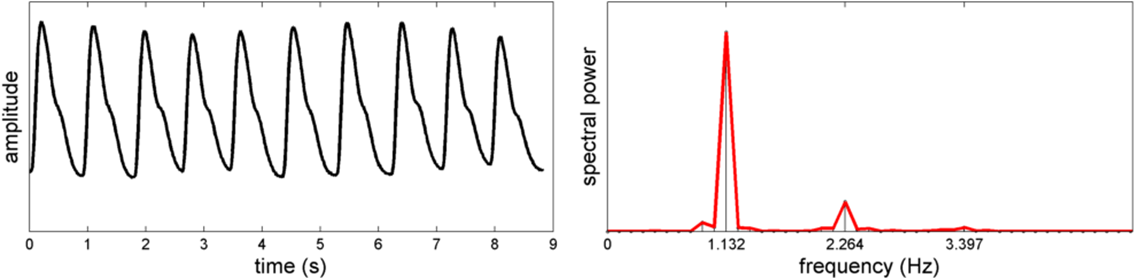}
\caption{Quasi-periodic nature of a typical blood pulse waveform signal. The periodicity and dicrotic characteristics of the waveform result in predominantly harmonic frequencies in the power spectral density.}
\label{fig:spectral}
\end{figure}

Ideally, $p(\vec{z}|X)$ should be a function of the SNR of the estimated temporal signal, since this provides information about the signal fidelity. However, SNR requires knowing the true signal, which is unknown at the time of acquisition. A proxy metric for estimating SNR should thus be computed using prior knowledge of blood pulse waveform characteristics. A spectral-spatial model is proposed based on the following two observations, which can be leveraged as prior information in the Bayesian framework presented above:
\begin{itemize}
  \item \textbf{Spectral:} Clean blood pulse waveforms are quasi-periodic, and are primarily composed of a weighted sum of a small set of sinusoidal signals (see Figure~\ref{fig:spectral}).
  \item \textbf{Spatial:} Non-homogeneous skin areas exhibit high variability due to anatomical non-uniformity (e.g., boundary, skin fold, hair).
\end{itemize}
For motivation, Figure~\ref{fig:spectral} shows a typical power spectral density of a clean blood pulse waveform. The spectral energy is compact, and is primarily composed of two harmonic frequencies. This indicates the quasi-periodic nature of the blood pulse waveform, and provides rationale for the spectral model.

In order to compute spectral properties, the normalized 0-DC spectral power distribution for spatial region $i$ was computed:
\begin{equation}
  \Gamma_i(f) = \frac{|F_i(f)|^2}{\int |F_i(f)|^2 df}
  \label{eq:gamma}
\end{equation}
where $F_i(f)$ are the complex-valued Fourier transform frequency coefficients for $x_i(t)-\bar{x}_i(t)$. The normalized spectral power (i.e., $\int \Gamma_i(f) df=1$) was used to model the relative AC pulsatile amplitude in the unit-less blood pulse waveforms. 

The quasi-periodic blood pulse waveform is dominated by the fundamental frequency corresponding to the heart rate and the first harmonic (see Figure~
\ref{fig:spectral}). To quantify this property, the spectral power exhibited by the fundamental frequency and first harmonic was computed:
\begin{equation}
  h_i = \int_{f^*-\Delta f}^{f^*+\Delta f} \Gamma_i(f) df + \int_{2f^*-\Delta f}^{2f^*+\Delta f} \Gamma_i(f) df
\end{equation}
where $f^*=\argmax_f{\Gamma_i(f)}$, and $\Delta f$ is the spectral window's half-width. We used $\Delta f = 0.2~Hz$. $h_i$ was set to 0 for signals whose fundamental frequency was outside of the physiologically realistic heart rate range. The final ``harmonic prior'' was computed as:
\begin{equation}
  w_i^{\text{harm}} = \exp \left( \frac{-(1-h_i)^2}{\alpha_h} \right)
\end{equation}
where $\alpha_h$ is a tuning parameter. An inverse exponential was used to emphasize small values of $(1-h_i)$ (i.e., strong harmonic contributions).

To quantify noise exhibited by the quasi-periodic waveform, the maximum spectral power response outside of the fundamental heart rate range was found:
\begin{equation}
  q_i = \max_f \left\{ 1-\int_{f^*-\Delta f}^{f^*+\Delta f} \Gamma_i(f) df\right\}
\end{equation}
The final ``noise prior'' was computed as:
\begin{equation}
  w_i^{\text{nmag}} = \exp \left( \frac{-q_i^2}{\alpha_q} \right)
\end{equation}
where $\alpha_q$ is a tuning parameter. An inverse exponential model was used to emphasize small values of $q_i$ (i.e., low noise).

Local anatomical variations may corrupt any pulsatile signals exhibited by underlying vessels (e.g., hair, skin fold, shadow ridge), or may not contain a pulsatile components at all (e.g., clothing, naris, eyelid). In order to estimate the anatomical uniformity at a given location, the image gradient was computed. In particular, given an image scene $\Lambda$ whose individual regions are $\vec{x}_i$, the ``spatial prior'' was computed as:
\begin{equation}
  w_i^{\text{spat}}= \exp \left( \frac{-\nabla \Lambda^2}{\alpha_l} \right)
\end{equation}
where $\nabla \Lambda$ is the gradient of image $\Lambda$. An inverse exponential model was used to emphasize small values of $\nabla \Lambda$ (i.e., homogenous areas).

The individual priors for region $i$ were combined to form the final region spectral-spatial probabilistic prior:
\begin{equation}
  W_i = \inf \left\{ \prod_k w_i[k] \mid N_i \right\}
\end{equation}
where $w_i = \{w_i^{\text{harm}}, w_i^{\text{nmag}}, w_i^{\text{spat}} \}$, and $N_i$ is the neighbourhood around region $i$. Here, a regional first order statistic constraint was imposed on the priors in order to further enforce spatial cohesion. Substituting this into Equation~\ref{eq:whist} produces the estimate of the posterior distribution $\hat{p}(\vec{z}|X)$.

\section{Results}
\subsection{Setup}
Data were collected across 24~participants of varying age (9--60~years, $(\mu \pm \sigma) = 28.7 \pm 12.4$) and body compositions (fat\% $30.0 \pm 7.9$, muscle\% $40.4 \pm 5.3$, BMI $25.5 \pm 5.2$~kg$\cdot$m$^{-2}$). Participants assumed a supine position throughout the study. A coded hemodynamic imaging (CHI) system was positioned facing down at the participant at a distance of 1.5~m, comprising a monochromatic camera with NIR sensitivity (Point Grey GS3-U3-41C6NIR) and a diffuse halogen illumination source (Lowel Rifa eX 44). To capture deep tissue penetration using NIR wavelengths, and to minimize the effects of visible environmental illumination (e.g., flicker), an 850--1000~nm optical bandpass filter was mounted in front of the camera lens. A video of each participant was recorded at 60~fps, with 16~ms exposure time. The frames were downsampled using $6 \times 6$ blockwise averaging. The ground truth PPG waveform was synchronously recorded using the Easy Pulse photoplethysmography finger cuff~\cite{easypulse}.

We compared our method, henceforth called \textbf{FusionPPG}, with DistancePPG~\cite{kumar2015} and ``FaceMeanPPG'', where the face is tracked and the signal is extracted through framewise spatial averaging. This method is commonly used in similar studies~\cite{sun2011,poh2010,mcduff2014}. Many pulse extraction methods rely on processing individual color channels~\cite{poh2010,mcduff2014,xu2014,dehaan2014}, and were therefore infeasible for this study (and infeasible in low-light settings, such as sleep studies). For our implementation of FaceMeanPPG, we spatially averaged the area identified by Viola-Jones face tracker~\cite{poh2010}. In its original implementation, DistancePPG requires estimating the true heart rate based on an averaging approach similar to FaceMeanPPG~\cite{kumar2015}. To generate optimal comparison results, DistancePPG was provided with the ground-truth heart rate (rather than their estimation method, which was found to fail in some cases).

In order to evaluate and compare signal fidelity between methods, normalized spectral entropy ($H$) and Pearson's linear correlation coefficient ($\rho$) were computed for each extracted signal:
\begin{equation}
  H(\hat{\vec{z}})=-\sum_{k=0}^{N-1} Z[k] \cdot \log Z[k]
\end{equation}
where $Z$ is the normalized spectral power for $\vec{z}$ according to Equation~\ref{eq:gamma}, and
\begin{equation}
  \rho(\hat{\vec{z}},\vec{y}) = \frac{|\sigma_{\hat{\vec{z}},\vec{y}}|}{|\sigma_{\hat{\vec{z}}}\sigma_{\vec{y}}|}
\end{equation}
where $\sigma_{\hat{\vec{z}}},\sigma_{\vec{y}}$ are the standard deviation of the extracted signal and ground-truth signal respectively, and $\sigma_{\hat{\vec{z}},\vec{y}}$ is the covariance between the two signals. To account for pulse time differences between the neck/head and finger, the maximum forward-sliding cross-correlation value within a short temporal window was used.

The heart rate of a blood pulse waveform signal was computed in the temporal domain using an autocorrelation scheme for increased temporal resolution~\cite{wander2014}. Specifically, each waveform was resampled at 200~Hz using cubic spline interpolation, and autocorrelation peaks were detected and used to estimate heart rate:
\begin{equation}
  \widehat{HR} = 60\frac{F_s}{\Delta t}
\end{equation}
where $F_s$ is the sampling rate, and $\Delta t$ is the time shift yielding peak autocorrelation response. Hyperparameter optimization was performed to find optimal tuning parameters using a grid search method with the following performance metric:
\begin{equation}
  \frac{\sum_{k \in \eta} \hat{Z}_k}{1-\sum_{k \in \eta} \hat{Z}_k}
\end{equation}
where $\hat{Z_k}$ is the $k^{\text{th}}$ Fourier coefficient of the estimated signal $\hat{z}$, and $\eta$ is the set of coefficients pertaining to the fundamental frequency and first harmonic of the $\hat{z}$. When choosing the optimal hyperparameters, signals exhibiting physiologically unrealistic heart rates were excluded.

One participant's data were removed due to erroneous ground-truth waveform readings. The study was approved by a University of Waterloo Research Ethics committee.

\subsection{Data Analysis}
\begin{figure}
\includegraphics[width=\textwidth]{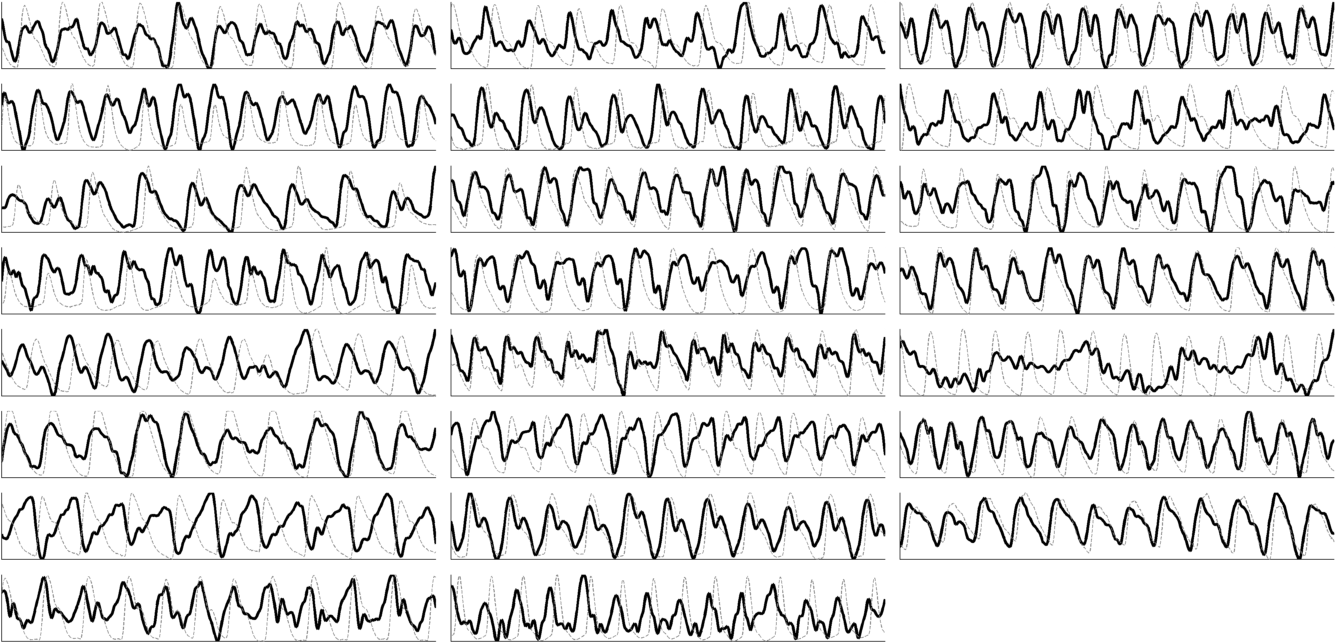}
\caption{Signals extracted from all 23~participants using the proposed FusionPPG method (black), plotted against to the ground-truth FingerPPG waveform (gray, dotted).}
\label{fig:results_allsigs}
\end{figure}

Figure~\ref{fig:results_allsigs} shows the signals extracted using the proposed fusion method compared to the ground-truth finger waveform. The waveforms exhibited high temporal fidelity, and were highly correlated to the ground-truth waveforms. The foot of each blood pulse waveform can be observed, signifying the precise time of the start of ventricular contraction. The method failed on one participant due to high fat content (42.3\%).

\begin{figure}
\centering
\includegraphics[width=\textwidth]{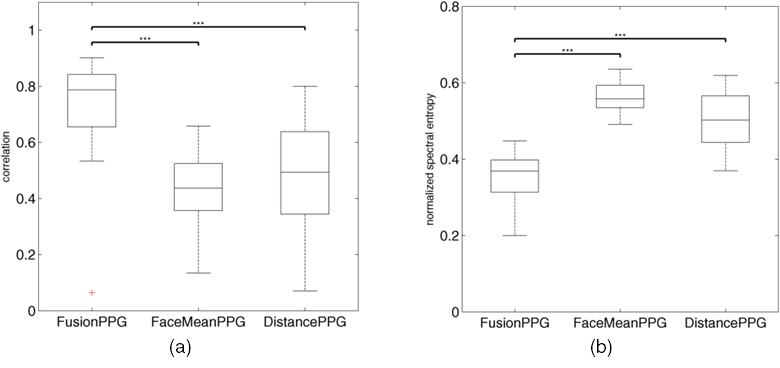}
\caption{Box plot comparison of the correlation (a) and normalized spectral entropy (b) between the signals extracted using the proposed (FusionPPG) and the two comparison (FaceMeanPPG, DistancePPG) methods. FusionPPG exhibited significantly higher correlation and significantly lower spectral entropy (i.e., higher spectral compactness) compared to FaceMeanPPG and DistancePPG. ($^{***}$statistically significant difference, $p<0.001$)}
\label{fig:significance}
\end{figure}

FusionPPG outperformed both comparison methods on the sample dataset. Figure~\ref{fig:significance} compares the box plot of the proposed and comparison methods using correlation (higher is better) and normalized spectral entropy (lower is better). FusionPPG attained statistically significantly higher correlation to the ground-truth waveform than FaceMeanPPG ($p<0.001$) and DistancePPG ($p<0.001$), signifying signals with higher temporal fidelity. FusionPPG also attained statistically significantly lower normalized spectral entropy than FaceMeanPPG ($p<0.001$) and DistancePPG ($p<0.001$), signifying more compact frequency components, consistent with the quasi-periodic nature of a true blood pulse waveform. DistancePPG attained higher correlation and lower entropy than FaceMeanPPG, consistent with previous findings~\cite{kumar2015}.

\begin{figure}
\centering
\includegraphics{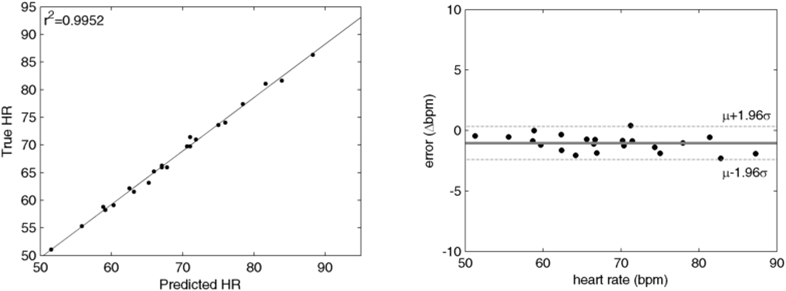}
\caption{Correlation and Bland-Altman plots of the predicted heart rates using the extracted blood pulse waveform signal. The predicted heart rates were highly correlated to the ground-truth heart rate ($r^2=0.9952$), and were in tight agreement ($\mu=-1.0~\text{bpm},\sigma=0.70~\text{bpm}$). The outlier was omitted due to failed signal extraction.}
\label{fig:blandaltman}
\end{figure}

FusionPPG was able to precisely estimate heart rate from the extracted waveforms. Figure~\ref{fig:blandaltman} shows the correlation and Bland-Altman plots showing FusionPPG's ability to extract precise and accurate heart rate. The predicted heart rates were highly correlated to the ground-truth heart rate ($r^2=0.9952$), and were in tight agreement, with low mean error ($\mu=-1.0~\text{bpm}$) and low variance ($\sigma=0.70~\text{bpm}$). The data were well represented within two standard deviations from the mean. The outlier was omitted from this analysis due to failed signal extraction.

\begin{figure}
\centering
\includegraphics[width=\textwidth]{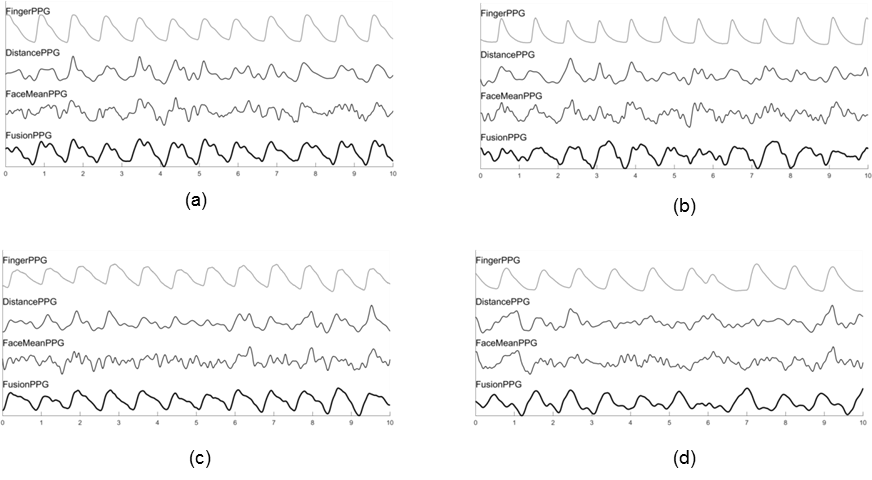}
\caption{Extracted waveforms from the proposed and comparison methods across four participants. The selected waveforms were those that exhibited the strongest correlation from DistancePPG~(a), FaceMeanPPG~(b), FusionPPG~(c), and a participant with arrhythmia~(d). FusionPPG was able to extract strong waveforms across all participants, and was the only method where the arrhythmia in (d) was visually apparent (at $t=6$~s).}
\label{fig:compplot}
\end{figure}

Figure~\ref{fig:compplot} compares the extracted waveforms from four participants using the three methods to the ground-truth waveform. The strongest waveforms (i.e., highest correlation) from DistancePPG, FaceMeanPPG, and FusionPPG were shown. An important characteristic is the foot of the waveform, which signifies the start of ventricular contraction. This foot was observed in each case, whereas it was not easily discernible in either DistancePPG or FaceMeanPPG due to the effects of averaging, resulting perhaps in a strong fundamental frequency which can predict heart rate, but is affected by spurious irrelevant frequencies that corrupt the waveform shape.

Figure~\ref{fig:compplot}(d) shows a participant that experienced a cardiac arrhythmia. An irregular cardiac contraction was observed at $t=6~\text{s}$, resulting in a delayed contraction. Such cases cannot be observed using standard heart rate analysis in the frequency domain. However, the irregular heartbeat and delayed follow-up contraction was observed in FusionPPG's waveform, whereas it was not apparent in FaceMeanPPG or DistancePPG. This demonstrates the important of temporal signal fidelity to assess irregular cardiac events that deviate from typical waveforms.

\section{Discussion}
In this study, the most pulsatile areas were often found in the neck region. The neck contains important vascular pathways, including the carotid arteries, which are major vessels that are closely connected to the heart and are close to the surface compared to other major arteries in the body. Pulsatile information in the face is more subdued, since the small arteries and arterioles are found further down the vascular tree. Thus, many existing methods that extract signals from the face may be at a disadvantage, and miss the rich information present in the neck.

The extraction method failed on the participant who had the highest fat \% of the sample. Skin folds and thick tissue layers contributed to the inability to extract a signal with any method. Many existing studies do not provide participant body composition, which is an important parameter when assessing signal strength.

An important characteristic that is not often discussed in PPGI studies is its ability to extract and assess abnormal waveforms (e.g., arrhythmia). In order to be useful as a health monitoring system, a PPGI system must not directly or indirectly assume normal waveforms. This was apparent in the arrhythmia case (Figure~\ref{fig:compplot}(d)), where only FusionPPG's waveform was able to temporally convey an abnormal cardiac event. During validation, emphasis should be placed on detected abnormal as well as normal waveforms.

Traditional heart rate variability is assessed through the RR peak intervals using an electrocardiogram. However, similar timing differences can be observed and quantified using the blood pulse waveform. An important part of this waveform is the blood pulse foot, which is the minimum point just prior to inflection due to the oncoming blood pulse. The blood pulse is ejected from the heart due to left ventricular contraction, which is directly controlled by the electrical signals governing the heart mechanics. The timing difference between the ECG's R peak and the PPG's foot is the pulse transit time. Thus, timing differences between the blood pulse feet indicate timing differences in the heart~\cite{schafer2013}.

Many existing methods, including DistancePPG and FaceMeanPPG, require tracking and/or segmenting the individual's face. However, it may be beneficial to assess pulsatility in areas other than the face (e.g., arm, hand, leg, foot). These methods will fail at this task since no face will be detected. In contrast, FusionPPG does not make any \apriori assumptions about anatomical locations, and may therefore be used to assess pulsatility at other anatomical locations in future work.

\section{Conclusions}
We have proposed a probabilistic signal fusion framework, FusionPPG, for extracting a blood pulse waveform signal from a scene using physiologically derived prior information. This was accomplished by posing the problem as a Bayesian problem and modeling the posterior distribution as a novel probabilistic pulsatility model that incorporated spectral and spatial priors derived from blood pulse waveform physiology. Results showed signals with strong temporal fidelity. The proposed method's improvement over the comparison methods was statistically significant ($p<0.001$) by assessing correlation and normalized spectral entropy. The FusionPPG waveform was the only one where cardiac arrhythmia was identifiable in the temporal domain. The model has presented such that it allows for future extensions based on this general theoretical framework

\section*{Funding}
This work was supported by the Natural Sciences and Engineering Research Council (NSERC) of Canada, AGE-WELL NCE Inc., and the Canada Research Chairs program.

\bibliography{bib}   % bibliography data in report.bib
\bibliographystyle{spiejour}

%\section{Acknowledgments}

%%%%% Biographies of authors %%%%%

\end{document}